# Implicit Tubular Surface Generation Guided by Centerline

Haoyin Zhou, James K. Min, and Guanglei Xiong, *Member, IEEE*

*Abstract*—Most machine learning-based coronary artery segmentation methods represent the vascular lumen surface in an implicit way by the centerline and the associated lumen radii, which makes the subsequent modeling process to generate a whole piece of watertight coronary artery tree model difficult. To solve this problem, in this paper, we propose a modeling method with the learning-based segmentation results by (1) considering mesh vertices as physical particles and using interaction force model and particle expansion model to generate uniformly distributed point cloud on the implicit lumen surface and; (2) doing incremental Delaunay-based triangulation. Our method has the advantage of being able to consider the complex shape of the coronary artery tree as a whole piece; hence no extra stitching or intersection removal algorithm is needed to generate a watertight model. Experiment results demonstrate that our method is capable of generating high quality mesh model which is highly consistent with the given implicit vascular lumen surface, with an average error of 0.08 mm.

## I. INTRODUCTION

Coronary artery disease (CAD) is the leading cause of death in the world. Computational fluid dynamics (CFD) applied to coronary computed tomography angiography (CTA) now allows for noninvasive calculation of coronary flow and pressure without additional medication or imaging [1] and has been shown to be effective for the diagnosis of CAD. A patient-specific geometric mesh model of coronary arteries segmented from CTA imaging is a prerequisite to perform CFD analysis of blood flow [2] [3]. The current methods to obtain a patient-specific coronary arteries model from CTA imaging typically involve the following three steps: the first step is coronary arteries centerline extraction [4] and using the centerline as a guide; the second step is vascular lumen segmentation [5]; and finally, the last step is generating mesh model of the vascular lumen according to the segmentation results. In this paper we mainly focus on the modeling procedure, which should ideally be fully-automated to reduce burden, improve reproducibility, and decrease the analysis time over the manual analysis. Most importantly, the generated mesh model should not only be consistent with the segmentation results but also be watertight, of high quality and has no mesh intersection.

The vascular lumen segmentation methods are mainly classified into voxel-based methods and machine learning-based methods. Conventional voxel-based segmentation methods are useful for the delineation of vascular geometry and the subsequent modeling procedure can be performed by applying marching cubes [6] or level set [7]. This kind of voxel-based methods is capable of yielding high quality meshes and constructing watertight model by handling bifurcation geometry; however, they are unable to incorporate expert knowledge, which limits the performance when compared to manually labeled ground truths.

On the other hand, machine learning-based methods have been proposed for the segmentation of cardiac structures with the advantages of not only increasing the segmentation speed but also learning from manual annotations. Typically, they assumed vessels to be cylindrical objects and detected vessel lumen boundaries along radial rays sampled uniformly from the centerline on the cross-section plane [8] [9] [10], as shown in Fig. 1. This will lead to an implicit and structured representing way of the segmented vascular lumen surface, which is defined as tubes along the centerline and the associate lumen radii radiated along the sampling directions [11] [12] [13]. This representing way makes it easy and straightforward to model each individual vessel segments respectively. However, for CFD analysis where a watertight and no-intersection mesh model is required, the subsequent lumen model construction becomes difficult with this representing way. The first cause is the bifurcation geometry is not appropriated to be represented in the form of centerline and associated lumen radii. Hence an extra bifurcation stitching method is required to constitute the whole piece of coronary arteries tree. Secondly, at some high curvature areas, the directions of the sampling cross-sections at the centerline points need to be chosen carefully, or they may lead to mesh intersections hence an extra mesh intersection removal method is needed. Therefore, to our knowledge, most previous learning-based methods assumed a loose combination of tubular structures and did not watertight model the coronary artery tree as a whole piece.

This paragraph of the first footnote will contain the date on which you submitted your paper for review. It will also contain support information, including sponsor and financial support acknowledgment.

This work was supported by the National Institutes of Health via grants #R21HL132277

Haoyin Zhou, James K. Min, and Guanglei Xiong are with the Dalio Institute of Cardiovascular Imaging and Department of Radiology, Weill Cornell Medical College, New York, NY, U.S.A. E-mail: [haz2011, jkm2001, gux2003] @med.cornell.edu

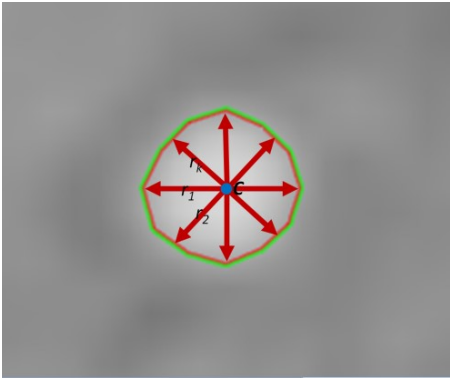

Fig. 1. Cross-section of a vessel. Implicit modeling of coronary lumen by a list of radial vectors with lumen radii $r_1, \cdots r_k$ from the centerline point $C$.

In this paper, we propose a novel method for the construction of a watertight coronary arteries tree model from the given implicit vascular lumen surface obtained by machine learning-based segmentation. This method has mainly two advantages: (1) it considers the vascular tree model as a whole piece hence no extra stitching method is required, and; (2) it is capable of constructing high quality meshes from the given centerline and lumen radii, even at bifurcations. Our approach is comprised of the following two steps: the first step is generating uniform point cloud on the implicit lumen surface by locking the points on the implicit surface with the intermolecular forces (IMFs) model and scattering points with the interaction force model and expansion model; and then, an incremental Delaunay triangulation method is proposed to generated triangle meshes from the point cloud.

This paper is organized as follows: Section II will briefly introduce the centerline extraction and machine learning-based segmentation, whose results are used as the input of the following model construction methods. Section III will describe the detail of the uniform point cloud generation method and Section IV will describe the incremental Delaunay triangulation methods. Then we will give the experiments results in Section V and finally conclude with a discussion and future work in Section VI.

*A. Related Works*

Coronary artery segmentation from CTA imaging has studied for decades [14]. Most segmentation methods firstly detect the centerline [4] and then segment vascular lumen guided by detected centerline [5].

Most centerline detection methods start with either heuristics-based [12, 15] or learning-based [16] vessel enhancement filtering. By assuming vessels as a set of tubes, heuristics-based approaches are effective in highlighting tubular structures. In contrast, learning-based methods do not assume the vessel shape as a priori, but make use of the geometric and information embedded in manual annotations. Based on the enhanced image, Yang et al proposed a data-driven centerline tracing method [16]. Zheng et al demonstrated a more robust method to trace main trunks of the coronary arteries by using a prior shape model [17].

Lumen segmentation methods are mainly classified into two types: voxel-based methods and machine learning-based methods. Voxel-based methods are widely used. For example, Antiga et al generated patient-specific vessel meshes for CFD analysis by using level set [18]. Wang et al integrated the level set method in a framework by iteratively refining centerlines and detecting vessel boundaries using level-sets [19]. Another work by Shahzad et al segmented lumen by combining graph cuts and kernel regression [20] in order to accurately detect stenosis. Several open source libraries were developed, such as VMTK [21] and TubeTK [22], which provide API functions for implementing level set and some vessel segmentation researches were done based on them or some commercial software [23] [24]. To improve accuracy and reduce the mesh density, Santis et al proposed to use hexahedral meshes [25]. The above voxel-based methods are capable of handling bifurcation geometry and construct seamless model; however, they are relatively slow and unable to incorporate expert knowledge, which limits the performance when compared to manually labeled ground truths.

Machine learning based segmentation method is faster and able to learn from manual annotations. Lugauer et al used the probabilistic boosting tree (PBT) to detect vessel lumen boundary by training the PBT classifier from manually labeled ground truths [26]. However, bifurcations are difficult to be modeled with leaning-based segmentation; hence extra methods are required for bifurcation modeling. Auricchio *at el* proposed a method to generate the bifurcation mesh by first obtaining the interfaces between each branches [26]; and then decoupling branches by these interfaces and modeling each branch as a tubular object. Another work proposed by Antiga deeply studied the geometrical relationship of the branches at bifurcations. Based on this, they sub-divided a bifurcation into four parts - three branches and one triangular-based prism and proposed a method to stitch the surface of each branch [27]. This type of methods are efficient for generating the bifurcation model with three branches, but are difficult to handle more branches.

II. SEGMENTATION METHOD

The flowchart of our system is given in Fig. 2. The segmentation method firstly applied to provide data for the subsequent modeling process. In this section, we will briefly introduce our vascular segmentation method from CTA imaging. In segmentation procedure of our system, the centerlines are firstly extracted from CT imaging by using the Frangi filter [28] and a thinning algorithm [29]. By resampling a given centerline (dense in our case), a list of uniformly distributed nodes is generated as centerline nodes, $c_1,...,c_l$, and a smoother centerline is obtained by finding a spline curve interpolating them. We define a local coordinate frame $[t,u,v]$ at each node $c$ using a rotation minimization technique, where $t$ is along the tangent direction of the centerline and $[u,v]$ spans a 2D plane on the cross-section. Then, the vascular lumen surface is implicitly represented as a list of lumen radii, $r_1,...r_k$, radiated from every centerline point $c$ along $k$ radial vectors sampled uniformly on the 2D $[u,v]$ planes. The lumen radii $r_1,...r_k$ are obtained by using machine learning-based lumen boundary segmentation, which will be introduced later. To make the surfaces of the lumen smooth, two types of interpolation-based refinements are used. One is in the longitudinal direction by refining the centerline nodes and calculating corresponding lumen radii for the new nodes. The other is in the circumferential direction by refining the lumen distances.

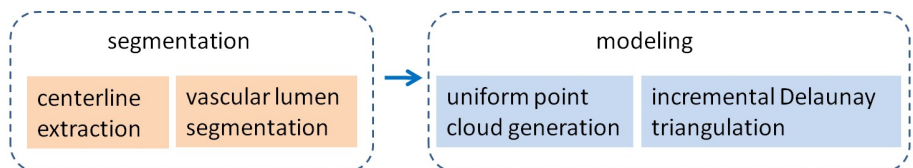

Fig. 2. The flowchart of our system.

For lumen boundary radii segmentation, we trained a boosting-based [30] classifier to segment the vascular lumen. For each center point $c$, we search for the optimal lumen distances $r_k$ separately by using the boosting-based segmentation classifier, which is trained on a manually annotated data set including 113 CTA images and over 15 million sample points. The features we used to train this classifier form a $25 \times 1$ vector:

$$\begin{bmatrix} I & I^2 & |\mathbf{g}| & |\mathbf{g}|^2 & \mathbf{g} \cdot \mathbf{n} & \cdots \end{bmatrix}_{25 \times 1}^T \quad (1)$$

where $I$ is CT image intensity, $\mathbf{g}$ is CT image gradient, $\mathbf{n}$ is the normal direction of the model surface. The feature vector includes the five components (see (1)) calculated at a point itself and other four neighborhood points along the normal direction. The $|\mathbf{g}|$ and $|\mathbf{g}|^2$ components suggest the likelihood of being at edges. The $\mathbf{g} \cdot \mathbf{n}$ component suggests the level of alignment of the CT gradient direction with the model surface normal direction. At the testing phase when the vessel model is previously unknown, $\mathbf{n}$ is initialized to be the direction of the associated radial vector.

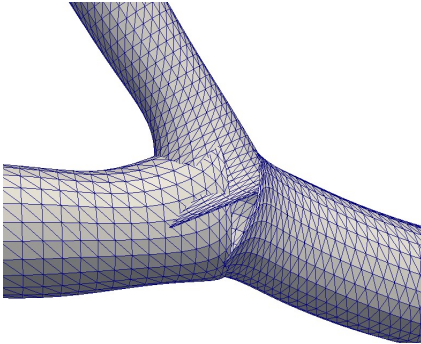

Fig. 3. The data structured from machine learning-based segmentation makes it difficult to model the bifurcation part.

This segmentation method links centerlines with structured vascular lumen radii, which makes it easy and straightforward to model each individual vessel segments respectively. However, as shown in Fig. 3, it is not able to model the bifurcation part directly. Hence an extra stitching method, such as convex hull generation, is required to obtain a watertight coronary artery model. This kind of methods will make the modeling procedure complex because they need to process the bifurcations differently and then combine the bifurcation parts and tubular vessel parts. It will be more graceful and more easily implemented if all vessel segments can be considered as a whole piece.

## III. UNIFORM POINT CLOUD GENERATION

Obtaining uniformly distributed vertices is essential for generating a high quality mesh model of the vascular lumen surface. Because of the complex shape, it is difficult to generate a point cloud distributed evenly throughout the implicit surface, which is represented in the form of centerline and the associated lumen radii. To solve this problem, we propose a physical model in which the mesh vertices are considered as physical particles, each particle receives forces from the centerline and neighboring particles and moves accordingly. Eventually, a uniform distribution will be obtained.

The physical model used in our algorithm describes the relationship between particle motion and forces. For each particle $\mathbf{p}_i$, we have

$$d\mathbf{p}_i = \left( \mu \mathbf{F}_{cl,i} + \sum_{j \in \Omega_i} \mathbf{F}_{i,j} \right) / m \quad , \quad (2)$$

where $\mathbf{F}_{cl,i}$ is the interaction force between the centerline and particle $\mathbf{p}_i$; $\mathbf{F}_{i,j}$ is the interaction force between particle $\mathbf{p}_i$ and $\mathbf{p}_j$ ($i \neq j$); $\Omega_i$ is the set of neighboring particles of $\mathbf{p}_i$; $m$ is the particle mass which is the same for all particles; $\mu$ is a coefficient which controls the relative weights between $\mathbf{F}_{cl,i}$ and $\mathbf{F}_{i,j}$.

### A. Centerline Force $\mathbf{F}_{cl,i}$

The centerline force $\mathbf{F}_{cl,i}$ is introduced to lock particles on the implicit lumen surface, whose direction will be kept along the particle normal direction $\mathbf{n}_i$:

$$\mathbf{F}_{cl,i} = \left| \mathbf{F}_{cl,i} \right| \cdot \mathbf{n}_i \quad (3)$$

where $|\cdot|$ suggests the magnitude of a vector; $\mathbf{n}_i$ is the unite normal direction vector of $\mathbf{p}_i$.

Denote $d_{i,t}$ as the Euclidean distance between particle $\mathbf{p}_i$ and its closest centerline point $\mathbf{p}_t$. The associated vascular lumen radius along the direction $\mathbf{p}_t \to \mathbf{p}_i$ is denoted as $R_t$. To estimate the magnitude of $\mathbf{F}_{cl,i}$, the simplified intermolecular forces (IMFs) model [31] is employed:

$$\left| \mathbf{F}_{cl,i} \right| = \frac{1}{\left( \frac{\alpha d_{i,t}}{R_t} + (1-\alpha) \right)^6} - \frac{1}{\left( \frac{\alpha d_{i,t}}{R_t} + (1-\alpha) \right)^3} \quad (4)$$

where $\alpha$ is a coefficient and we choose $\alpha = 0.5$.

(4) defines the magnitude of centerline-particle force $\mathbf{F}_{cl,i}$. As shown in Fig. 4, $|\mathbf{F}_{cl,i}| = 0$ when $d_{i,t} = R_t$. When $d_{i,t} < R_t$, $|\mathbf{F}_{cl,i}| > 0$ suggests the centerline will repel $\mathbf{p}_i$; when $d_{i,t} > R_t$, $|\mathbf{F}_{cl,i}| > 0$ suggests the centerline will attract $\mathbf{p}_i$. In this way, $|\mathbf{F}_{cl,i}|$ locks the particles at the lumen surface which is implicitly represented by the centerline and the associated radii. When $\mathbf{p}_i$

is far away from the centerline, the centerline force $|\mathbf{F}_{cl,i}|$ will gradually reduce to zero.

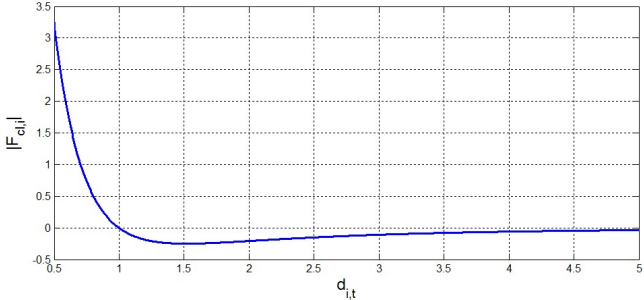

Fig. 4. Relationship between $d_{i,t}$ and $|\mathbf{F}_{cl,i}|$ according to (4). ($R_t = 1.0$);

### B. Particle Normal Direction $\mathbf{n}_i$

As shown in Fig. 5(a), the estimation of the normal direction $\mathbf{n}_i$ is easy and straightforward at the tubular vascular part:

$$\mathbf{n}_i = \mathbf{p}_i - \mathbf{p}_t \quad (5)$$
$$\mathbf{n}_i = normalize(\mathbf{n}_i)$$

where $\mathbf{p}_t$ is the closest centerline point to $\mathbf{p}_i$.

However, at bifurcation areas where the curvature of the surrounding centerline is large, it is not appropriate to calculate $\mathbf{n}_i$ according to (5), as shown in Fig. 5 (b). Because $\mathbf{n}_i$, which is the direction of $\mathbf{F}_{cl,i}$, will dramatically change even with a small displacement, which will eventually lead to a low quality points distribution in this local area.

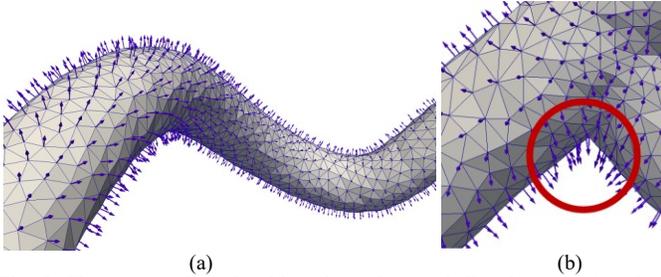

(a) (b)
Fig. 5. The generated mesh with estimated normal directions $\mathbf{n}_i$ (a) at low curvature areas, normal direction can be obtained by finding the closest centerline point; (b) at high curvature areas (shown in red circle), the normal direction estimation should take into account all surrounding centerlines.

The algorithm to detect and estimate the normal direction $\mathbf{n}_i$ of a particle $\mathbf{p}_i$ at a high curvature area near a bifurcation is given in Tab. 1. Based on geometric analysis, this algorithm is able to generate smoothly varying normal direction $\mathbf{n}_i$ with respect to the location of $\mathbf{p}_i$. This is important because (1) the smoothly varying $\mathbf{n}_i$ is important to make the points relocating smoothly, which is a prerequisite for achieving a uniform distribution, and; (2) for the Delaunay triangulation, a proper projection direction is needed to avoid mesh intersection.

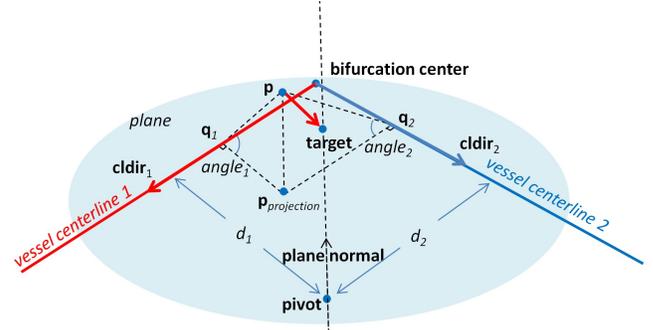

Fig. 6. Schematic diagram for demonstrating the geometric relationships used in the algorithm in Tab. 1.

TABLE I
Algorithm: get normal $\mathbf{n}$ for a point at high curvature area near a bifurcation.

| | |
|---|---|
| 1 | **if** $|\mathbf{p} - \mathbf{bifurcation\ center}| < 1.5 \times$ neighboring lumen radius<br>    **if** $\mathbf{cldir}_1 \cdot \mathbf{cldir}_2 > 0$    // **cldir** is the centerline direction radiated from bifurcation center<br>       $\mathbf{dir}_{mid} = (\mathbf{cldir}_1 + \mathbf{cldir}_2)/2$;<br>       **If** the unit vector of $(\mathbf{p} - \mathbf{bifurcation\ center})$ is close to $\mathbf{dir}_{mid}$<br>           $\mathbf{p}$ is a high curvature point<br>       **end if**<br>    **end if**<br>**end if** |
| 2 | **if** $\mathbf{p}$ is not a high curvature point, **pause** |
| 3 | Find point **pivot** on the plane defined by **bifurcation center** and *centerlines*, denote $d_1$ and $d_2$ as the distances between **pivot** to centerlines. |
| 4 | Project point $\mathbf{p}$ to the plane and get $\mathbf{p}_{projection}$, project $\mathbf{p}$ to centerlines and obtain $\mathbf{q}_1$ and $\mathbf{q}_2$. |
| 5 | Set $angle_1 = \angle \mathbf{pq}_1\mathbf{p}_{projection}, angle_2 = \angle \mathbf{pq}_2\mathbf{p}_{projection}$ |
| 6 | Calculate $\lambda_1 = d_1 \tan(angle_1), \lambda_2 = d_2 \tan(angle_2)$ |
| 7 | Calculate $\lambda = (\lambda_1 + \lambda_2)/2$, calculate **target** = **pivot** + $\lambda$ (**plane normal**) |
| 8 | $\mathbf{n} = \mathbf{target} - \mathbf{p}; \mathbf{n} = normalize(\mathbf{n})$; |

### C. Particles Repelling Force $\mathbf{F}_{i,j}$

Some natural or social phenomenon shows entities have a tendency of becoming uniformly distributed. For instance, assuming other conditions are similar, a crowed city tends to drive people out of it while an empty city attracts people. Gradually, the populations of all cities become closed.

In our algorithm, we employ this kind of phenomenon that entities repelling each other to avoid a too high or too low density. We use a simple physical model of considering each particle as a balloon which is gradually inflated. The balloons will repel each other when contacting. In this model we need to solve two problems: (1) how to calculate the repelling force; and (2) how to inflate the balloon.

The particle-particle interaction force $\mathbf{F}_{i,j}$ is always perpendicular to the normal direction $\mathbf{n}_i$, which is used to scatter the particles and will eventually lead to a uniform distribution. For two particles $\mathbf{p}_i$ and $\mathbf{p}_j$, the repelling forces between them have the same magnitude and opposite directions. The direction of $\mathbf{F}_{ij}$ is determined by first obtaining the ray defined by $\mathbf{p}_i$ and $\mathbf{p}_j$, and then project this ray according to the normal direction $\mathbf{n}_i$. The magnitude of repelling force between them is:

$$|\mathbf{F}_{i,j}| = \begin{cases} 0 & \text{if } d_{i,j} > 0.5(r_i + r_j) \\ \dfrac{1}{\left(\dfrac{\alpha d_{i,j}}{r_i + r_j} + (1-\alpha)\right)^6} - \dfrac{1}{\left(\dfrac{\alpha d_{i,j}}{r_i + r_j} + (1-\alpha)\right)^3} & \text{else} \end{cases} \quad (6)$$

where $r_i$ and $r_j$ are the balloon radii of $\mathbf{p}_i$ and $\mathbf{p}_j$ respectively.

The resultant force $\sum_{j \in \Omega_i} \mathbf{F}_{i,j}$ together with $\mathbf{F}_{cl,i}$ will determine the motion of $\mathbf{p}_i$, where $\Omega_i$ is the set of neighboring points to $\mathbf{p}_i$:

$$\Omega_i = \{p_j \mid d_{i,j} < 0.5(r_i + r_j)\} \quad (7)$$

In our algorithm, each balloon $\mathbf{p}_i$ will gradually dilate by imaginary inflation, but the particle interaction forces $\{\mathbf{F}_{i,j}\}, j \in \Omega_i$ may compress the balloon.

The particle interaction force applied on $\mathbf{p}_i$ can be divided into two parts (see (8)): the first part is involved in the motion of $\mathbf{p}_i$ and the second part compresses the balloon. We have:

$$\sum_{j \in \Omega_i} |\mathbf{F}_{i,j}| = \left|\sum_{j \in \Omega_i} \mathbf{F}_{i,j}\right| + |\mathbf{F}_{compress,i}| \quad (8)$$

Considering an ideal hexagonal close-packing, which is the best uniform sampling pattern and all balloons stop changing its radius, we have:

$$\left|\mathbf{F}^{ideal}_{compress,i}\right| = \frac{6}{\left(\frac{0.5 \times 0.5(r_i + r_j)}{r_i + r_j} + 0.5\right)^6} - \frac{6}{\left(\frac{0.5 \times 0.5(r_i + r_j)}{r_i + r_j} + 0.5\right)^3} \quad (9)$$
$$= 19.4897$$

Hence, $r_i$ will be updated according to the compress force:

$$dr_i = -\ln(1.0 + \eta(|\mathbf{F}_{compress,i}| - |\mathbf{F}^{ideal}_{compress,i}|)) \quad (10)$$

$$\begin{cases} \text{if } dr_i > dr_{max}, dr_i = dr_{max} \\ \text{if } dr_i < dr_{min}, dr_i = dr_{min} \end{cases} \quad (11)$$

where $\eta$ is a coefficient suggesting the hardness of the balloon.

According to (10), when $|\mathbf{F}_{compress,i}| < |\mathbf{F}^{ideal}_{compress,i}|$, the radius $r_i$ will become larger, and according to (6), $\mathbf{p}_i$ will have larger interaction forces with neighboring points which tends to make $|\mathbf{F}_{compress,i}|$ larger. Similarly, when $|\mathbf{F}_{compress,i}| > |\mathbf{F}^{ideal}_{compress,i}|$, our algorithm tends to make $|\mathbf{F}_{compress,i}|$ smaller. This forms a negative feedback system hence the all variables in our algorithm, including points distribution, will become stable.

Fig. 7 intuitively demonstrates the point cloud redistribution process. After calculation, points have larger radii and perform a uniform distribution.

## IV. INCREMENTAL DELAUNAY TRIANGULATION

After obtaining a uniform point cloud on the implicit vascular lumen surface, in this section, an incremental Delaunay triangulation algorithm is proposed to triangulate the point cloud.

Because of the tubular structure and the uniform distribution, each point has connections only with its neighbors. Hence, it is appropriate to do the triangulation locally rather than globally.

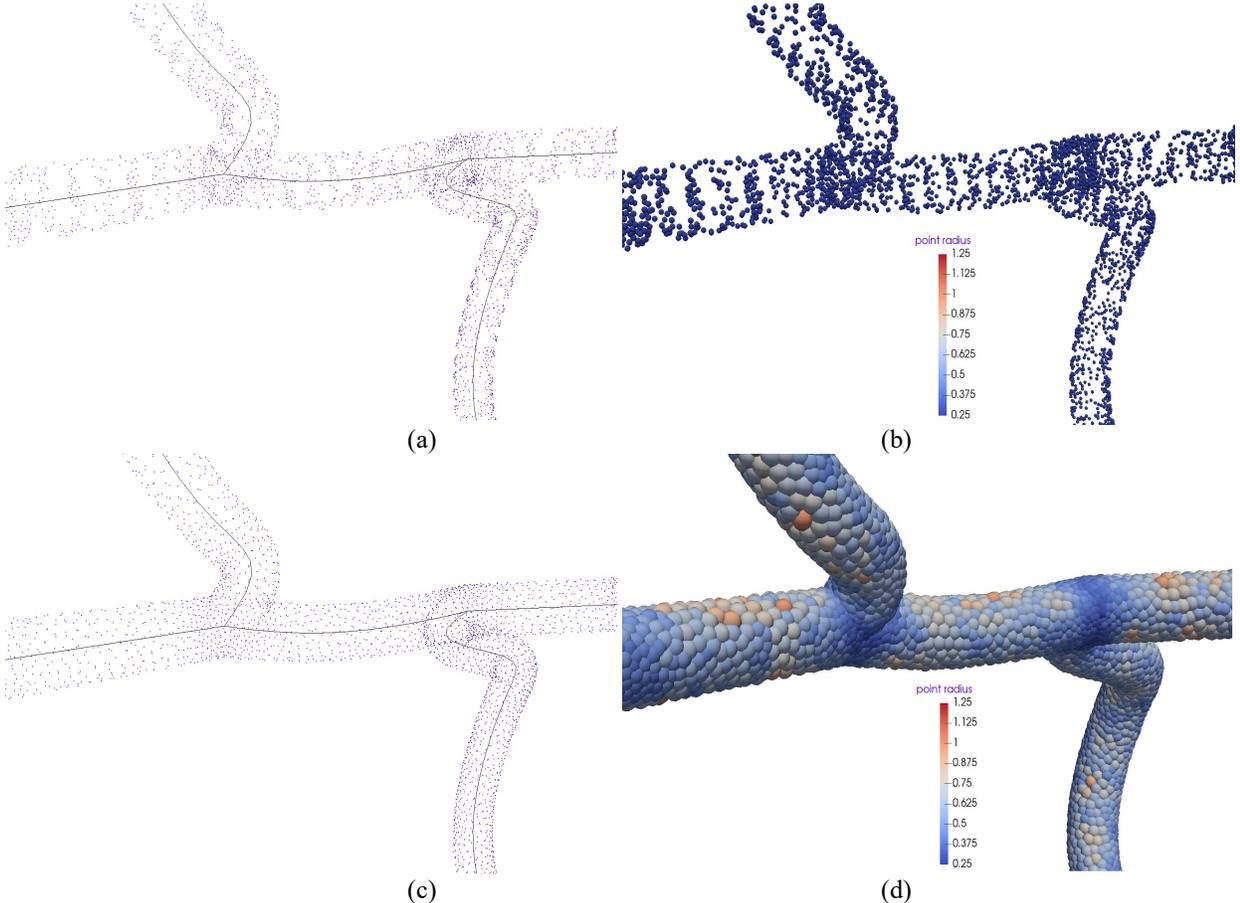

Fig. 7. (a) At the beginning of our algorithm the point distribution is not uniform; (b) the initial point's radii are small. (c) After our algorithm uniform distribution is achieved; (d) the point's radii expand so they have interaction forces between each other.

The proposed Delaunay triangulation approach inserts points and its related triangles into the mesh model one by one. For each point only its neighboring areas are considered in the triangulation procedure.

Another reason for not using a globally triangulation method is that it is difficult to find a projection way to project all 3D points to a 2D plane without causing intersections, because the curvature of the vascular lumen surface is high and the shapes at bifurcations are complex.

TABLE II
The incremental Delaunay triangulation algorithm

| | |
|---|---|
| Input | point cloud: $P = \{p_i\}$, $i = 1,2,...N$; |
| Output | Mesh $M = \{P, T\}$; |
| | $T$ is the triangle meshes set, $T = \{triangles\}$; |
| Initial: | Insert at least one initial points to vector *Queue*; |
| | *index* = 0; |
| 1 | **repeat** |
| 2 | $p_i$ = *Queue.at(index)*, $n_i$ = normal of $p_i$; |
| 3 | Define neighboring points set $\Omega_i = \{p_j | p_j$ is close to $p_i\}$; |
| 4 | Project all $p_j \in \Omega_i$ along $n_i$; |
| | Obtain 2D points set $\Omega_{2D,i} = \{p_{2D,j} | p_j \in \Omega_i\}$; |
| 5 | Find $T_{exist}$ = {triangles| all three vertices $\in \Omega_i$}; |
| 6 | **for** all $p_{2D,k1}, p_{2D,k2} \in \Omega_{2D,i}$, $k_1, k_2 \neq i$ |
| 7 | *flag = true*; // this is a flag indicating whether triangle |
| | // $\{p_i, p_{k1}, p_{k2}\}$ is a Delaunay triangle or not |
| 8 | **if** triangle $\{p_{2D,i}, p_{2D,k1}, p_{2D,k2}\}$ has intersection with $T_{exist}$, |
| | **then** *flag = false; continue*; |
| 9 | generate the minimum circumscribed circle of |
| | $\{p_{2D,i}, p_{2D,k1}, p_{2D,k2}\}$ and get center **c** and radius $r$ |
| 10 | **for** any $p_{2D,t} \in \Omega_{2D,i}$, $t \neq i, k_1$ and $k_2$ |
| 11 | **if** $|p_{2D,t} - c| < r$ && $\{p_{2D,t}, p_{2D,k1}, p_{2D,k2}\}$, |
| | $\{p_{2D,t}, p_{2D,i}, p_{2D,k1}\}$ or $\{p_{2D,t}, p_{2D,i}, p_{2D,k2}\}$ |
| | has no intersection with $T_{exist}$ |
| | **then** *flag = false; continue*; |
| 12 | **end for** |
| 13 | **if** *flag = true* |
| | **then** insert new triangle $\{p_i, p_{k1}, p_{k2}\}$ into $T$; |
| | insert $p_{k1}, p_{k2}$ into *Queue*; |
| | **end for** |
| 14 | *index = index* +1; |
| 15 | **when** *index* is not the end of *Queue* |

For each growing step of our incremental construction Delaunay triangulation algorithm, it is a constrained Delaunay triangulation process with constrictions from existing triangles. The projection direction for $p_i$ is its normal direction $n_i$, which is discussed in Section III.B.

Step 8 and 11 reflect the differences between constrained Delaunay and regular Delaunay. In step 8, if this candidate triangle $\{p_i, p_{k1}, p_{k2}\}$ will cause intersection, then this triangle should not be inserted into the mesh model. Step 11 separates different situations as shown in Fig. 7. In Fig. 7, $p_t$ is inside of the minimum circumscribed circle, suggesting the three triangles $\{p_t, p_{k1}, p_{k2}\}$, $\{p_t, p_i, p_{k1}\}$ and $\{p_t, p_i, p_{k2}\}$ has a larger minimum angle than triangle $\{p_i, p_{k1}, p_{k2}\}$ without considering the existing triangle. In Fig. 8 (a), however, the existing triangle makes these three triangles unavailable, while in Fig. 8 (b), triangle $\{p_i, p_{k1}, p_t\}$ is available and is a better choice than $\{p_i, p_{k1}, p_{k2}\}$.

With this triangulation algorithm, triangle meshes are generated from the generated uniform point cloud, as shown in Fig. 9.

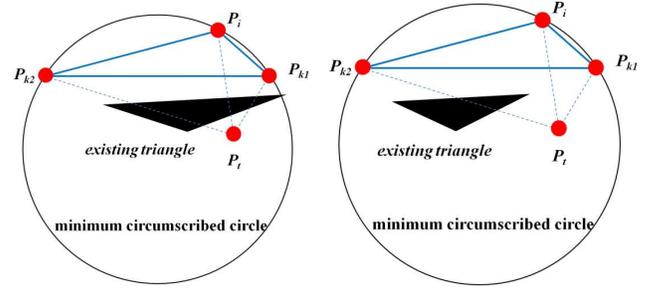

Fig. 8. Constrained Delaunay triangulation. (a) Although $p_t$ is inside of the minimum circumscribed circle, it shouldn't be taken into account because of the existing triangle. (b) the existing triangle cannot block the connections between $p_i$, $p_{k1}$ and $p_t$.

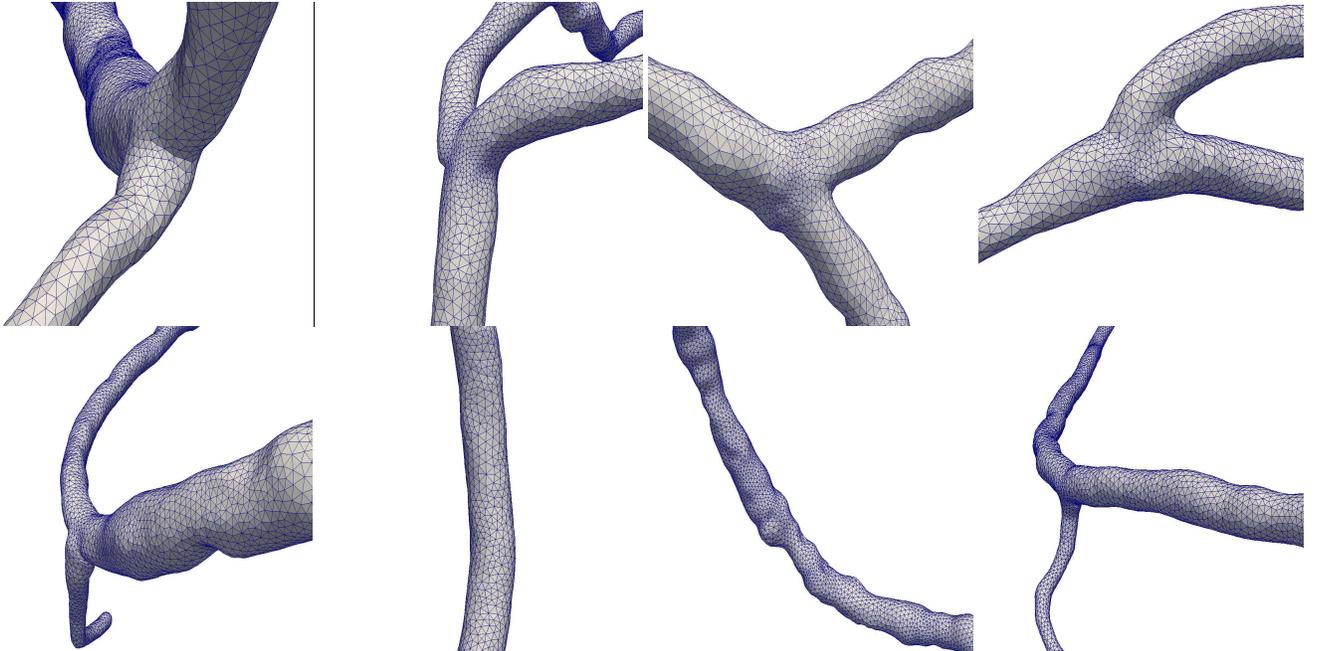

Fig.9. Demonstrations of the generated mesh, including bifurcation parts and tubular parts.

## V. EXPERIMENTS

Our algorithms have been implemented in C++ code running on a 2.60 GHz Intel Core i7 processer. We evaluate the proposed algorithms on real CTA imaging data from 25 patients.

The modeling algorithms are the main concern of this paper, which require the centerline and lumen radii as input. In this evaluation, we first used our previous work, the SmartCoronary software [11], to obtain vessel centerlines and lumen radii from the CTA images. SmartCoronary is capable of generating structured vascular lumen mesh model autonomously by firstly extracting vessel centerlines and then segmenting lumen radii with a machine learning-based method. Besides, it also allows users to adjust the segmentation results manually to correct possible mistakes.

With the obtained centerlines and lumen radii, we evaluate the performance of our modeling approach, which mainly have two steps: (1) uniform point cloud generation and; (2) incremental Delaunay triangulation. The initial points are randomly generated nearby all centerline points. For each centerline point, the number of the generated points is approximately proportional to its lumen radius.

### A. Runtime

Both algorithms used in the modeling process do the calculation by processing points one by one. When processing a point, only its neighboring points or triangles are taken into consideration. When the neighboring area is defined by a Euclidean distance-based thresholding method, the number of points inside the neighboring area is determined by the density of the point cloud, which is proportional to the total number of points $N$. Hence, both algorithms have a time complexity of $O(N^2)$. However, it is more appropriate to consider a smaller area with a higher the point cloud density. In our code, we consider the neighboring area of a point by obtaining the closest $M = 25$ points. In this way, both algorithms reduce the time complexity from $O(N^2)$ to $O(N)$.

In the following experiments we will evaluate the runtime to verify the above analysis. Firstly, we evaluate the runtime of the uniform point cloud generation algorithm. For this evaluation, the termination condition is set as:

$$\frac{1}{N}\sum_{i=1}^{N}\left|\mathbf{p}_i^{(t+1)}-\mathbf{p}_i^{(t)}\right|<0.05 \quad (12)$$

where ($t$) denotes the iteration number.

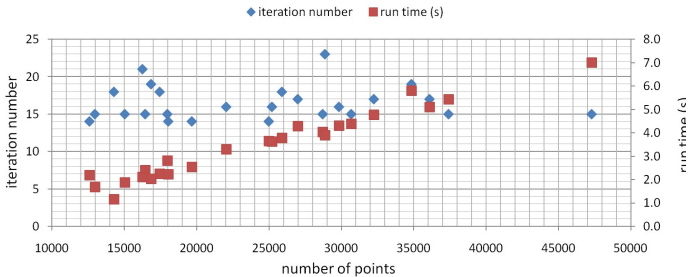

Fig. 10. Speed evaluation for the uniform point cloud generation algorithm. Runtime and the required iteration number on 25 patients' data with different number of model points.

Fig. 10 shows the runtime and the required number of iteration of our uniform point cloud generation algorithm on the 25 patients' data. It is clear demonstrated that the runtime has a linear relationship with the number of points, which is consistent with our theoretical analysis. As shown in Fig. 10, the algorithm needs around 20 iterations on each case suggesting it has high converge speed and stability. The average runtime for every 10,000 points is 1.404 s, which is fast.

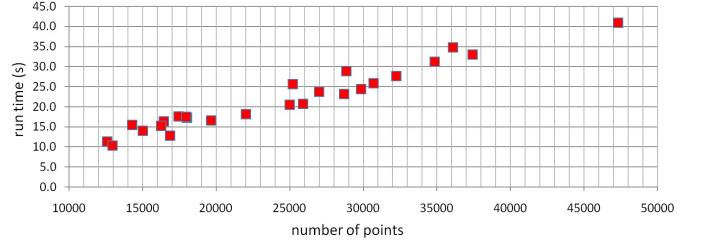

Fig. 11. Speed evaluation for the incremental Delaunay triangulation algorithm. Runtime on 25 patients' data.

Then we evaluate the runtime of the growing Delaunay triangulation algorithm. As shown in Fig. 11, approximately, the growing Delaunay triangulation algorithm also demonstrates a linear time complexity with respect to the number of model points. The average runtime for every 10,000 points is 8.989 s.

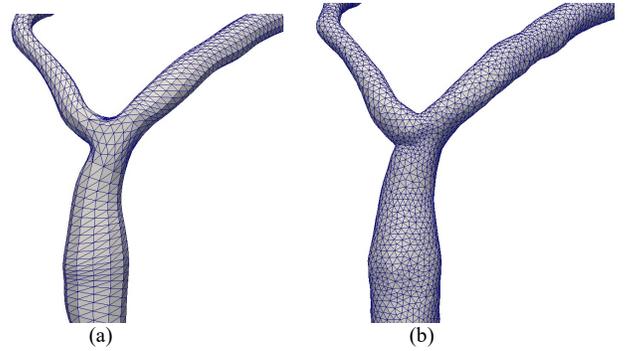

Fig. 12. (a) Structured mesh, with bifurcation mesh generated by convex hull and mesh smoothing. (b) The unstructured mesh obtained by the proposed modeling algorithms in this paper.

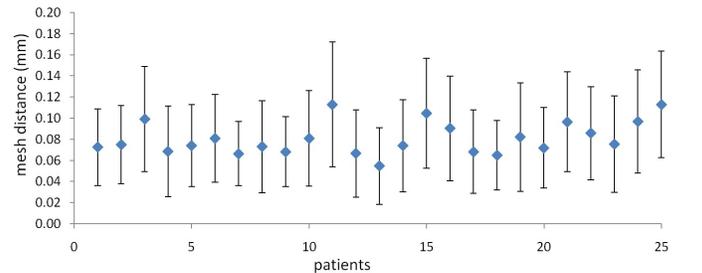

Fig. 13. Mesh distance comparison results between the structured and unstructured mesh on 25 patients' data.

### B. Accuracy

It is of crucial importance for the proposed modeling approach to reflect the given centerline and lumen radii accurately. To evaluate the modeling accuracy, we introduce the structured mesh generated directly from centerlines and lumen radii [11] as shown in Fig. 12, and the bifurcation mesh is obtained by firstly generating the convex hull and then smoothing mesh. Then we calculate the mesh distance between the generated structured and unstructured mesh. The comparison results are given in Fig. 13. The average error distance is 0.080 mm. This very small error distance suggesting the accuracy of our modeling approach is high.

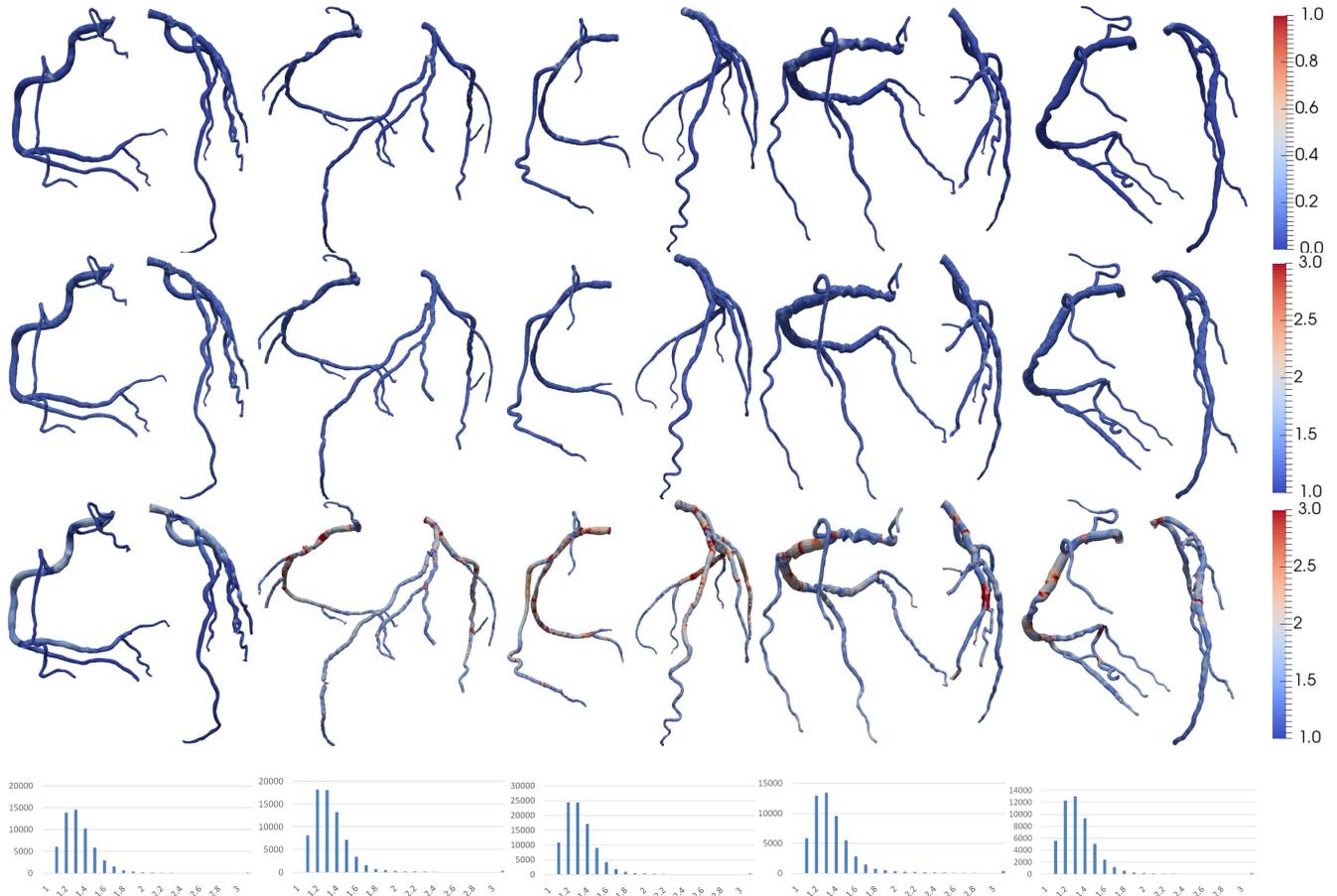

Fig. 15. Results demonstrations. The first row shows the mesh distance comparison results (mm). The second row shows the mesh quality evaluation results of the unstructured mesh generated by approaches proposed in this paper. The third row shows the mesh quality evaluation results of the structured mesh. The last row gives the histogram of mesh quality of the unstructured mesh.

## C. Mesh Quality

We employ *mesh quality* = (*longest edge*) / (*shortest edge*) as the metrics to evaluate the mesh quality. An ideal triangle mesh model should be composed of right triangles with *quality = 1* for all triangles. As shown in Fig. 14, our algorithms are able to generated stable high quality triangle meshes compared with the structured mesh model.

We demonstrate some of the experiment results in Fig. 15, including the evaluation results of the accuracy, the mesh quality of both unstructured and structured mesh, to demonstrate that the proposed algorithm is capable of handling the complex shape of the coronary arteries tree.

## VI. CONCLUSION AND FUTURE WORK

We propose a coronary arteries modeling method which is suitable for combining machine learning-based segmentation. It can be regarded as a bridge between learning-based segmentation and CFD analysis. Experiment results show that it is robust and can reflect the segmentation results accurately.

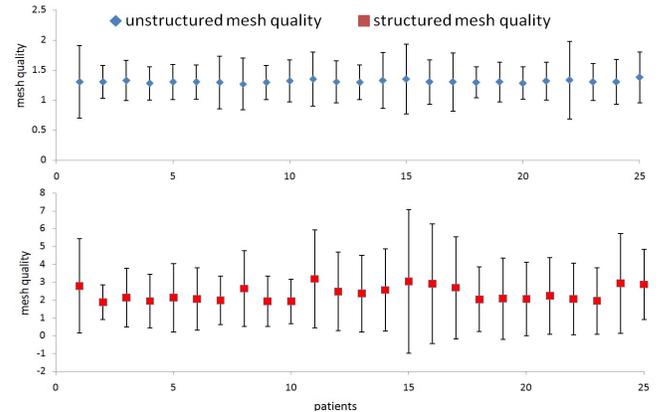

Fig. 14. Mesh quality evaluation results on the structured and unstructured mesh on 25 patients' data.

One advantage of this modeling method is it is able to generate unstructured meshes from given structured segmentation results without losing global topology and connectivity of vessel segments. Unstructured meshes are excellent for representation of complex surface details.

The uniform point cloud generation algorithm is concise and of elegance and can process the whole model as a whole piece. However, extra normal direction estimation algorithm is required at high curvature areas. Future work should continue on studying how to obtain robust normal directions.

In future work we should also continue on improving the point cloud generation algorithm to allow it to generate or remove points autonomous according to local point density. This will make the whole model to have a similar point density.

REFERENCES

[1] H. Kim, I. Vignon-Clementel, J. Coogan, C. Figueroa, K. Jansen, and C. Taylor, "Patient-specific modeling of blood flow and pressure in human coronary arteries," *Annals of biomedical engineering,* vol. 38, pp. 3195-3209, 2010.

[2] G. A. Holzapfel, J. J. Mulvihill, E. M. Cunnane, and M. T. Walsh, "Computational approaches for analyzing the mechanics of atherosclerotic plaques: a review," *Journal of biomechanics,* vol. 47, pp. 859-869, 2014.

[3] C. A. Taylor, T. A. Fonte, and J. K. Min, "Computational fluid dynamics applied to cardiac computed tomography for noninvasive quantification of fractional flow reserve: scientific basis," *Journal of the American College of Cardiology,* vol. 61, pp. 2233-2241, 2013.

[4] M. Schaap, C. T. Metz, T. van Walsum, A. G. van der Giessen, A. C. Weustink, N. R. Mollet*, et al.*, "Standardized evaluation methodology and reference database for evaluating coronary artery centerline extraction algorithms," *Medical image analysis,* vol. 13, pp. 701-714, 2009.

[5] H. Kirişli, M. Schaap, C. Metz, A. Dharampal, W. B. Meijboom, S. Papadopoulou*, et al.*, "Standardized evaluation framework for evaluating coronary artery stenosis detection, stenosis quantification and lumen segmentation algorithms in computed tomography angiography," *Medical Image Analysis,* vol. 17, pp. 859-876, 2013.

[6] W. E. Lorensen and H. E. Cline, "Marching cubes: A high resolution 3D surface construction algorithm," in *ACM siggraph computer graphics*, 1987, pp. 163-169.

[7] J. A. Sethian, *Level set methods and fast marching methods: evolving interfaces in computational geometry, fluid mechanics, computer vision, and materials science* vol. 3: Cambridge university press, 1999.

[8] F. Lugauer, J. Zhang, Y. Zheng, J. Hornegger, and B. M. Kelm, "Improving accuracy in coronary lumen segmentation via explicit calcium exclusion, learning-based ray detection and surface optimization," in *SPIE Medical Imaging*, 2014, pp. 90343U-90343U-10.

[9] O. Wink, W. J. Niessen, and M. A. Viergever, "Fast delineation and visualization of vessels in 3-D angiographic images," *Medical Imaging, IEEE Transactions on,* vol. 19, pp. 337-346, 2000.

[10] M. Hernández-Hoyos, A. Anwander, M. Orkisz, J.-P. Roux, P. Douek, and I. E. Magnin, "A Deformable Vessel Model with Single Point Initialization for Segmentation, Quantification, and Visualization of Blood Vessels in 3D MRA," in *Medical Image Computing and Computer-Assisted Intervention–MICCAI 2000*, 2000, pp. 735-745.

[11] G. Xiong, P. Sun, H. Zhou, S. Ha, B. Hartaigh, Q. Truong*, et al.*, "Comprehensive Modeling and Visualization of Cardiac Anatomy and Physiology from CT Imaging and Computer Simulations," 2016.

[12] M. A. Gülsün and H. Tek, "Robust vessel tree modeling," in *Medical Image Computing and Computer-Assisted Intervention–MICCAI 2008*, ed: Springer, 2008, pp. 602-611.

[13] R. Li and S. Ourselin, "Accurate curvilinear modelling for precise measurements of tubular structures," in *Digital Image Computing: Techniques and Applications; Proceedings of the VIIth Biennial Australian Pattern Recognition Society Conference, DICTA 2003*, 2003, p. 243.

[14] D. Lesage, E. D. Angelini, I. Bloch, and G. Funka-Lea, "A review of 3D vessel lumen segmentation techniques: Models, features and extraction schemes," *Medical image analysis,* vol. 13, pp. 819-845, 2009.

[15] C. Metz, M. Schaap, A. Weustink, N. Mollet, T. van Walsum, and W. Niessen, "Coronary centerline extraction from CT coronary angiography images using a minimum cost path approach," *Medical physics,* vol. 36, pp. 5568-5579, 2009.

[16] Y. Zheng, M. Loziczonek, B. Georgescu, S. K. Zhou, F. Vega-Higuera, and D. Comaniciu, "Machine learning based vesselness measurement for coronary artery segmentation in cardiac CT volumes," in *SPIE Medical Imaging*, 2011, pp. 79621K-79621K-12.

[17] Y. Zheng, H. Tek, and G. Funka-Lea, "Robust and accurate coronary artery centerline extraction in CTA by combining model-driven and data-driven approaches," in *Medical Image Computing and Computer-Assisted Intervention–MICCAI 2013*, ed: Springer, 2013, pp. 74-81.

[18] L. Antiga, B. Ene-Iordache, and A. Remuzzi, "Computational geometry for patient-specific reconstruction and meshing of blood vessels from MR and CT angiography," *Medical Imaging, IEEE Transactions on,* vol. 22, pp. 674-684, 2003.

[19] C. Wang, R. Moreno, and Ö. Smedby, "Vessel segmentation using implicit model-guided level sets," in *MICCAI Workshop" 3D Cardiovascular Imaging: a MICCAI segmentation Challenge", Nice France, 1st of October 2012.*, 2012.

[20] R. Shahzad, H. Kirişli, C. Metz, H. Tang, M. Schaap, L. van Vliet*, et al.*, "Automatic segmentation, detection and quantification of coronary artery stenoses on CTA," *The international journal of cardiovascular imaging,* vol. 29, pp. 1847-1859, 2013.

[21] L. Antiga and D. Steinman, "The vascular modeling toolkit," *URL: http://www. vmtk. org,* 2008.

[22] S. Aylward, D. Pace, A. Enquobahrie, M. McCormick, C. Mullins, C. Goodlett*, et al.*, "TubeTK, segmentation, registration, and analysis of tubular structures in images," *Kitware Inc., Clifton Park,* 2012.

[23] S. Morlacchi, S. G. Colleoni, R. Cárdenes, C. Chiastra, J. L. Diez, I. Larrabide*, et al.*, "Patient-specific simulations of stenting procedures in coronary bifurcations: two clinical cases," *Medical engineering & physics,* vol. 35, pp. 1272-1281, 2013.

[24] R. Cárdenes, J. L. Díez, I. Larrabide, H. Bogunović, and A. F. Frangi, "3D modeling of coronary artery bifurcations from CTA and conventional coronary angiography," in *Medical Image Computing and Computer-Assisted Intervention–MICCAI 2011*, ed: Springer, 2011, pp. 395-402.

[25] G. De Santis, M. De Beule, P. Segers, P. Verdonck, and B. Verhegghe, "Patient-specific computational haemodynamics: generation of structured and conformal hexahedral meshes from triangulated surfaces of vascular bifurcations," *Computer methods in biomechanics and biomedical engineering,* vol. 14, pp. 797-802, 2011.

[26] F. Auricchio, M. Conti, C. Ferrazzano, and G. A. Sgueglia, "A simple framework to generate 3D patient-specific model of coronary artery bifurcation from single-plane angiographic images," *Computers in biology and medicine,* vol. 44, pp. 97-109, 2014.

[27] L. Antiga, B. Ene-Iordache, L. Caverni, G. P. Cornalba, and A. Remuzzi, "Geometric reconstruction for computational mesh generation of arterial bifurcations from CT angiography," *Computerized Medical Imaging and Graphics,* vol. 26, pp. 227-235, 2002.

[28] A. F. Frangi, W. J. Niessen, K. L. Vincken, and M. A. Viergever, "Multiscale vessel enhancement filtering," in *Medical Image Computing and Computer-Assisted Intervention—MICCAI'98*, ed: Springer, 1998, pp. 130-137.

[29] J. R. Jensen, "Introductory digital image processing: a remote sensing perspective," Univ. of South Carolina, Columbus1986.

[30] R. E. Schapire and Y. Singer, "Improved boosting algorithms using confidence-rated predictions," *Machine learning,* vol. 37, pp. 297-336, 1999.

[31] I. Prigogine, A. Bellemans, and V. Mathot, *The molecular theory of solutions* vol. 4: North-Holland Amsterdam, 1957.